\documentclass{article} 
\usepackage{iclr2025_conference,times}

\usepackage{amsmath,amsfonts,bm}

\def\eqref#1{equation~\ref{#1}}

\def\1{\bm{1}}

\def\rmD{{\mathbf{D}}}
\def\rmE{{\mathbf{E}}}

\def\rmR{{\mathbf{R}}}

\DeclareMathAlphabet{\mathsfit}{\encodingdefault}{\sfdefault}{m}{sl}
\SetMathAlphabet{\mathsfit}{bold}{\encodingdefault}{\sfdefault}{bx}{n}

\makeatletter
\def\thanks#1{\protected@xdef\@thanks{\@thanks
        \protect\footnotetext{#1}}}
\makeatother

\usepackage{hyperref}       
\usepackage{url}            
\usepackage{booktabs}       
\usepackage{amsfonts}       
\usepackage{nicefrac}       
\usepackage{microtype}      
\usepackage{xcolor}         
\usepackage{multirow}
\usepackage{multicol}
\usepackage{graphicx}
\usepackage{enumitem}
\usepackage{amssymb}
\usepackage{bbding}
\usepackage{wrapfig}
\usepackage{subfigure}
\usepackage{capt-of}
\usepackage{authblk}
\usepackage{float} 

\usepackage{tipa}
\usepackage{pifont}
\usepackage{makecell}
\usepackage{utfsym}

\title{\centerline{VidTok} \vspace{-2mm} \leavevmode \centerline{A Versatile and Open-Source Video Tokenizer}}

\author{
\textbf{Anni Tang}$^{1,2}$, \textbf{Tianyu He}$^{\dag,1}$\thanks{$^{\dag}$Project lead.}, \textbf{Junliang Guo}$^1$, \textbf{Xinle Cheng}$^3$, \textbf{Li Song}$^2$, \textbf{Jiang Bian}$^1$}
\affil{
$^1$Microsoft Research, $^2$Shanghai Jiao Tong University, $^3$Peking University
}
\affil{
\url{https://github.com/microsoft/VidTok}
}

\iclrfinalcopy
\begin{document}

\maketitle

\begin{abstract}
Encoding video content into compact latent tokens has become a fundamental step in video generation and understanding, driven by the need to address the inherent redundancy in pixel-level representations. Consequently, there is a growing demand for high-performance, open-source video tokenizers as video-centric research gains prominence. We introduce VidTok, a versatile video tokenizer that delivers state-of-the-art performance in both continuous and discrete tokenizations. VidTok incorporates several key advancements over existing approaches: 1) model architecture such as convolutional layers and up/downsampling modules; 2) to address the training instability and codebook collapse commonly associated with conventional Vector Quantization (VQ), we integrate Finite Scalar Quantization (FSQ) into discrete video tokenization; 3) improved training strategies, including a two-stage training process and the use of reduced frame rates. By integrating these advancements, VidTok achieves substantial improvements over existing methods, demonstrating superior performance across multiple metrics, including PSNR, SSIM, LPIPS, and FVD, under standardized evaluation settings.
\end{abstract}

\begin{figure}[h]
  \centering
  \includegraphics[width=0.98\textwidth]{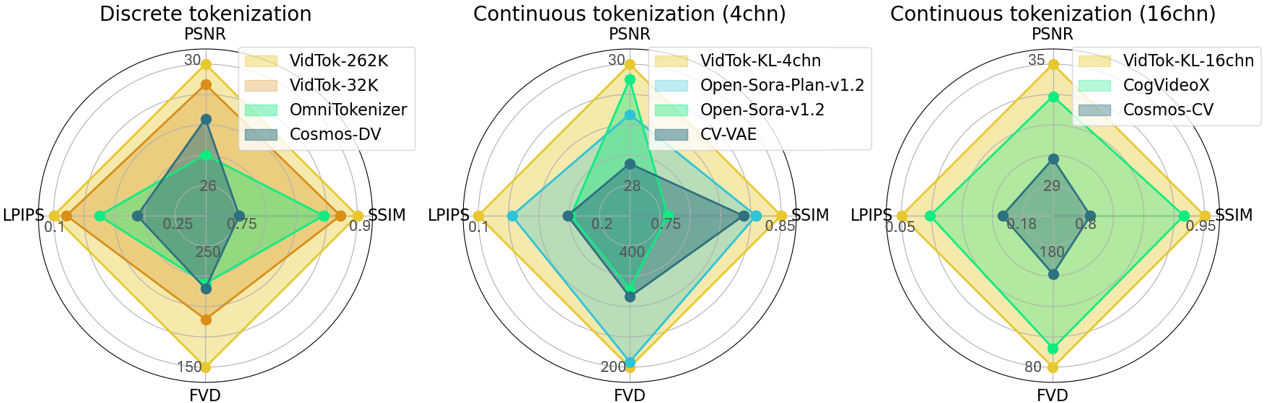}
  \caption{Illustration of the quantitative comparison of discrete and continuous tokenization performance across our VidTok model and state-of-the-art methods, evaluated using four metrics: PSNR, SSIM, LPIPS, and FVD. All performance metrics are obtained through experiments conducted under a consistent evaluation protocol to ensure fairness and comparability. Larger chart areas correspond to better performance across all metrics.}
  \label{fig:demo}
  \vspace{-0mm}
\end{figure}

\section{Introduction}
\label{sec:intro}

Visual generation and understanding have emerged as prominent research areas, driven by the capacity of visual data to offer immersive experiences~\citep{ho2022video,singer2023make,ho2022imagen,yu2023magvit,kondratyuk2024videopoet,yang2024cogvideox,bai2024uniedit,zhu2024compositional}, convey rich semantic information~\citep{li2023videochat,zhang2024video,liu2024visual}, and function as an interface for models to interact with the physical world~\citep{pmlr-v235-yang24z,yang2024learning,zhang2024videoicl,chen2024igor}. However, the high degree of redundancy inherent in pixel-level representations~\citep{sullivan2012overview} has led to a shift in modern methodologies. These approaches often employ visual tokenization techniques~\citep{rombach2022high,sora,kondratyuk2024videopoet,wu2024janus,team2024chameleon}, transforming raw visual data into compact latent tokens, which serve as a more efficient basis for tasks involving generation and understanding.

The adoption of visual tokenization has catalyzed extensive research on image tokenizers~\citep{rombach2022high,zheng2022movq,patil2024amused}, resulting in the development of several open-source tokenizers that serve as widely used tools to advance and streamline image-related research~\citep{repolatent,repomuse,repomagvit2}. However, comparable resources and tools remain largely absent in the domain of video. While it is possible to treat each frame of a video as an independent image and compress it using an image tokenizer, this approach overlooks temporal redundancies and consistency, resulting in latent tokens that are temporally redundant and potentially inconsistent across frames.

\begin{wrapfigure}{r}{0.6\textwidth}
  \begin{center}
    \includegraphics[width=0.58\textwidth]{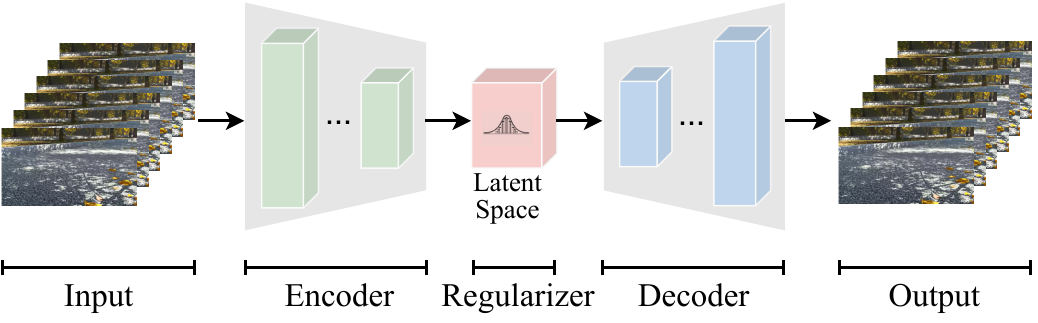}
  \end{center}
  \caption{An overview of video tokenizers.}
  \label{fig:overview}
\end{wrapfigure}

Recent efforts have sought to address this gap by introducing video tokenizers that incorporate temporal modeling. However, these approaches often fail to account for diverse use cases and exhibit limitations in performance. For instance, \citet{yang2024cogvideox} exclusively offers tokenizers with continuous tokens, while \citet{kondratyuk2024videopoet} demonstrates the effectiveness of discrete tokens but remains unavailable as an open-source tool. In this work, we introduce VidTok, a versatile and state-of-the-art video tokenizer designed to support both continuous and discrete tokenizations effectively. Our approach follows common architecture as illustrated in Fig.~\ref{fig:overview}, and incorporates several key advancements over existing solutions:

\begin{itemize}
    \item \textbf{Model architecture}. We handle spatial and temporal sampling separately, reducing computational complexity without sacrificing reconstruction quality. Specifically, we employ 2D convolutions in spatial up/downsampling modules and adopt an AlphaBlender operator in temporal up/downsampling modules, while the remaining parts still utilize 3D convolutions to perform information fusion.
    \item \textbf{Advanced quantization techniques}. To address the training instability and codebook collapse commonly associated with conventional Vector Quantization (VQ)~\citep{van2017neuralvqvae}, we propose the use of Finite Scalar Quantization (FSQ) in discrete video tokenization. By optimizing the implicit codebook directly, this approach substantially improves discrete tokenizers.
    \item \textbf{Improved training strategies}. To improve training efficiency, we employ a two-stage training strategy: initially pre-training the full model on low-resolution videos, followed by fine-tuning only the decoder on high-resolution videos. Furthermore, we observe that utilizing training data with reduced frame rates effectively improves the model's ability to represent motion dynamics.
\end{itemize}

Building upon the aforementioned advancements, we train VidTok on a large-scale video dataset and evaluate its performance on widely used benchmarks such as MCL-JCV~\citep{wang2016mcl} and a web video evaluation set. Experimental results reveal that VidTok outperforms previous models in both discrete and continuous tokenization, achieving superior results across all evaluated metrics, including PSNR, SSIM, LPIPS, and FVD.

\section{Related Works}
\label{sec:relatedwork}
\subsection{Discrete Video Tokenization}

Discrete tokenization maps input images to a latent space and quantizes the latent representations using a codebook of vectors by identifying the nearest codebook vector. Compared to continuous tokens, discrete tokens offer the advantage of mitigating error accumulation during the autoregressive generation process. Building on the foundation of discrete image tokenization~\citep{van2017neuralvqvae}, discrete video tokenization extends this approach to video data~\citep{yan2021videogpt,yu2024language,wang2024omnitokenizer,repocosmos}. It incorporates temporal modeling to effectively manage the temporal redundancies inherent in video sequences.

VideoGPT~\citep{yan2021videogpt} leverages VQ-VAE~\citep{van2017neuralvqvae} to learn downsampled discrete latent representations of raw video data through the use of 3D convolutions and axial self-attention. Subsequently, a GPT-like architecture is employed to autoregressively model these discrete latents, utilizing spatio-temporal position encodings. This approach produces video samples that are competitive with state-of-the-art GAN-based models for video generation.
MAGVIT-v2~\citep{yu2024language} observes that the generation performance initially improves but then deteriorates for larger vocabulary in VQ-VAE, and decreasing the code embedding dimension when increasing the vocabulary size facilitates learning over the distribution of a large vocabulary~\citep{yu2022vectorquantized}. Building on this insight, MAGVIT-v2 reduces the embedding dimension of the VQ-VAE codebook to zero and introduces Lookup-Free Quantization (LFQ), which eliminates the embedding lookup process. This approach improves both reconstruction and generation quality in language models as vocabulary size increases.
As a concurrent work, Cosmos-Tokenizer~\citep{repocosmos} utilizes Finite Scalar Quantization (FSQ)~\citep{mentzer2024finite} to achieve discrete tokenization, where each dimension is quantized to a small, fixed set of values.

In this work, we integrate several key advancements, including FSQ, to develop a state-of-the-art discrete video tokenizer. The proposed tokenizer is designed to facilitate a wide range of applications in video analysis, generation, and modeling, fostering further innovation in the field.

\subsection{Continuous Video Tokenization}

Compared to discrete tokenization, continuous tokenization~\citep{zhao2024cv,repoopensora,chen2024od,yang2024cogvideox,repocosmos} generally offers higher reconstruction fidelity~\citep{rombach2022high}. It is typically employed in conjunction with continuous space modeling techniques, such as diffusion models~\citep{ho2020denoising}, to enhance the quality and smoothness of generated outputs. For example, Latent Video Diffusion Models (LVDMs)~\citep{blattmann2023stable,guo2023animatediff,yang2024cogvideox,sora} efficiently and effectively generate video content by compressing visual data into continuous latent representation first and then operating on it with denoising techniques. A notable example of this approach is OpenAI’s Sora~\citep{sora}, which serves as a representative work in this domain.

CV-VAE~\citep{zhao2024cv} introduces a continuous video tokenizer designed to achieve spatio-temporal compression of videos, with a latent space that aligns with the latent space of existing image VAEs~\citep{rombach2022high} through its proposed latent space regularization method. Open-Sora~\citep{repoopensora} and Open-Sora-Plan~\citep{repoopensoraplan,chen2024od} are two open-source projects aimed at reproducing OpenAI’s Sora. Both projects offer continuous video tokenizers that effectively perform spatial and temporal compression. CogVideoX~\citep{yang2024cogvideox} introduces a continuous tokenizer that preserves a greater amount of information by maintaining a larger number of latent channels, resulting in enhanced reconstruction fidelity. More recently, Cosmos-Tokenizer~\citep{repocosmos} also provides continuous video tokenizers with various compression ratios.

The proposed VidTok builds upon the publicly available models mentioned above by incorporating several key advancements, with the goal of establishing a foundational tokenizer for video-related research.

\section{VidTok}
\label{sec:method}

In this section, we first introduce the general structure of the video tokenizer with detailed notations. From Sec.~\ref{sec:arch} to Sec.~\ref{sec:training_strategy}, we introduce the improved model architecture, the advanced quantization technique, and the improved training strategy respectively.

\subsection{Overview of Video Tokenizer}
\label{sec:overview}

To enhance efficiency, existing approaches for video generation and understanding often utilize video tokenizers (e.g., 3D VAEs~\citep{Kingma2014}) to convert raw visual data into compact latent tokens. As illustrated in Fig.~\ref{fig:overview}, these methods typically involve an encoder that compresses video data into compact latent tokens across both spatial and temporal dimensions, followed by a decoder that reconstructs the tokens back into pixel space. Depending on the scenario, latent tokens can be either continuous~\citep{zhao2024cv,yang2024cogvideox,sora} or discrete~\citep{yu2024language,wang2024omnitokenizer,repocosmos}, and the model architecture may be designed to operate in a causal~\citep{yu2024language} or non-causal~\citep{blattmann2023stable} manner. To enhance the model's capacity for generating novel data samples and to mitigate overfitting to the training dataset, it is essential to apply appropriate regularization within the latent space~\citep{Kingma2014,van2017neuralvqvae}.

Let $\rmE$ and $\rmD$ denote the encoder and the decoder of the video tokenizer respectively, $\rmR$ denote the regularizer applied in the latent space, $r_t$ and $r_s$ denote the temporal and spatial compression ratios respectively.
A video containing $N$ frames is denoted as $X = \{x_1, x_2, ..., x_N\} \in \mathbb{R} ^{N \times 3 \times H \times W}$, where $N=r_t*n$, $H=r_s*h$ and $W=r_s*w$. 
The workflow can be formulated as:
\begin{equation}
    \label{equ:1}
        Z = \rmR ( \rmE (X) ), \\ 
        \hat{X} = \rmD (Z)
\end{equation}
where $Z \in \mathbb{R} ^{n \times c \times h \times w}$ denotes the compressed latent representation and $\hat{X} \in \mathbb{R} ^{N \times 3 \times H \times W}$ denotes the reconstructed video.

In causal scenarios, the first frame is typically treated as an independent image for compression, enabling the visual tokenizer to function as both an image and video tokenizer~\citep{yu2024language}. At this point, a video $X \in \mathbb{R} ^{(N+1) \times 3 \times H \times W}$, which contains $N+1$ frames, is compressed into $Z \in \mathbb{R} ^{(n+1) \times c \times h \times w}$. Specifically, the first frame is compressed solely in the spatial dimension, while the subsequent frames undergo compression in both temporal and spatial dimensions.

\subsection{Model Architecture}
\label{sec:arch}

\begin{figure*}[t]
  \centering
  \includegraphics[width=0.95\linewidth]{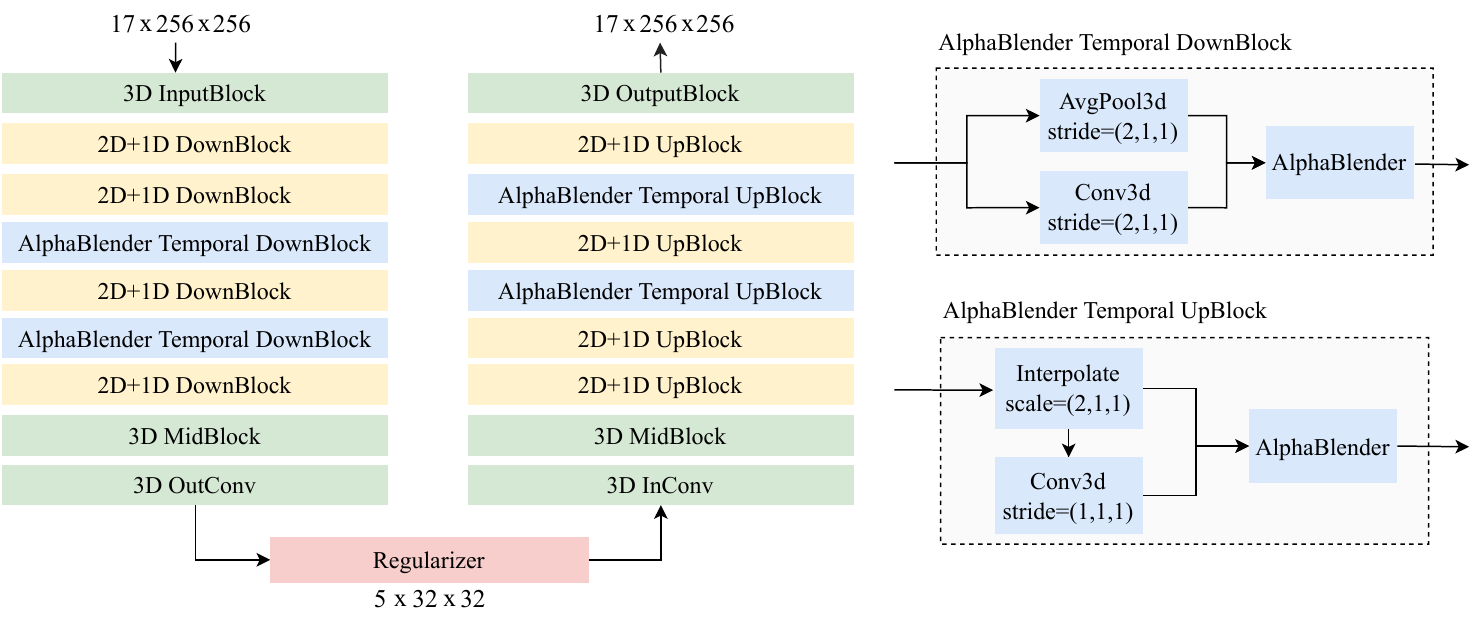}
  \caption{The improved model architecture. In the context of a causal setting, consider an input with dimensions $T \times H \times W = 17 \times 256 \times 256$. Assuming a temporal compression factor of $4$ and a spatial compression factor of $8$, the intermediate latent representation is reduced to dimensions $T \times H \times W = 5 \times 32 \times 32$.}
  \label{fig:arch}
\end{figure*}

In the existing literature, it is widely acknowledged that fully 3D architectures offer superior reconstruction quality, albeit at a high computational cost~\citep{chen2024od}. However, in this work, we demonstrate that substituting a portion of these 3D convolutions with a combination of 2D and 1D convolutions—effectively decoupling spatial and temporal sampling—can achieve comparable reconstruction quality while significantly reducing computational demands.

The detailed network architecture is illustrated in Fig.~\ref{fig:arch}. As shown, 2D convolutions are employed for spatial upsampling and downsampling modules, while an AlphaBlender operator is utilized in the temporal upsampling and downsampling modules. The remaining components, including the input/output layers and bottleneck layers, leverage 3D convolutions to facilitate information fusion. The specific structures of the temporal upsampling and downsampling modules are depicted on the right side of Fig.~\ref{fig:arch}. Additionally, layer normalization~\citep{lei2016layer} is incorporated throughout the architecture to enhance stability and performance. Experimental results, as summarized in Tab.~\ref{tab:abla_arch}, validate the effectiveness of the proposed architectural design.

\paragraph{AlphaBlender operator.}
Given a parameter $\alpha$ within the range $[0,1]$, the AlphaBlender operator performs the following operation to input $x_1$ and input $x_2$:
\begin{equation}
    \label{equ:alphablender}
        x = \alpha * x_1 + (1-\alpha) * x_2
\end{equation}
where $x$ is the result after blending, and $\alpha$ can be either learnable or a given hyperparameter~\citep{repoopensoraplan}. In this work, we adopt a pre-defined $\alpha=Sigmoid(0.2)$.

In causal cases, all 3D and 1D convolutions are configured to operate causally, ensuring that each frame has access only to historical information from preceding frames. For a given video $X \in \mathbb{R} ^{(N+1) \times 3 \times H \times W}$, the first frame is duplicated $r_t - 1$ times and inserted before the original first frame. After completing the full workflow, the process yields $(n+1)*r_t$ frames. By discarding the first $r_t - 1$ frames, the reconstructed video $\hat{X} \in \mathbb{R} ^{(n*r_t+1) \times 3 \times H \times W}$ is obtained.

\subsection{Finite Scalar Quantization}
\label{sec:fsq}

Variational AutoEncoders (VAEs)~\citep{Kingma2014} are a class of generative models that map each data point, such as an image, from a complex dataset into a continuous distribution within a latent space, rather than assigning it to a single deterministic point. Conversely, the decoder performs the inverse operation, mapping representations from the latent space back to the original input space. However, due to the increasing demand for discrete latent variables, Vector Quantised-Variational AutoEncoder (VQ-VAE)~\citep{van2017neuralvqvae} were introduced. Unlike standard VAEs, VQ-VAEs map inputs to a finite set of vectors (i.e., codebook), through a process known as vector quantization. This approach represents each input by the closest vector in the codebook, which is learned during training. By combining the generative capabilities of VAEs with the advantages of discrete representations, VQ-VAEs provide a robust framework for various machine learning applications, including data compression, representation learning, and generative modeling.

\begin{figure}[t]
  \centering
  \includegraphics[width=0.98\linewidth]{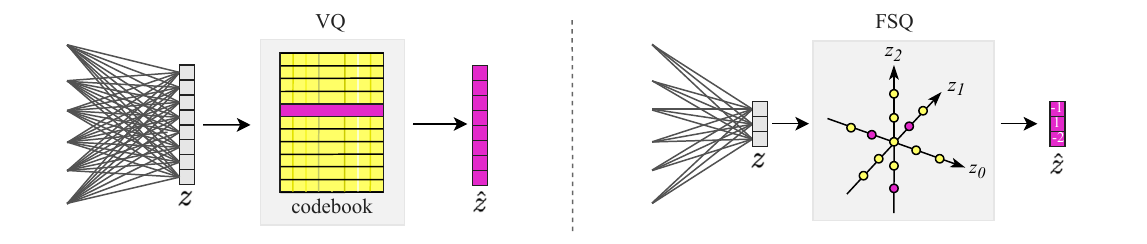}
  \caption{Left: Vector Quantization (VQ) employed in Vector Quantised-Variational AutoEncoder (VQ-VAE)~\citep{van2017neuralvqvae}. Right: Finite Scalar Quantization (FSQ)~\citep{mentzer2024finite} utilized in our model.}
  \vspace{-3mm}
  \label{fig:fsq}
\end{figure}

In this work, we employ Finite Scalar Quantization (FSQ)~\citep{mentzer2024finite} to generate discrete tokens. The central principle of FSQ is that each scalar entry in the latent representation is independently quantized to the nearest pre-defined scalar value through rounding. In contrast to Vector Quantization (VQ), FSQ eliminates the need for codebook learning, thereby improving training stability~\citep{mentzer2024finite,yu2024language}. The approach can be described as follows: Given a vector $z=(z_1, z_2, ..., z_d)$ with $d$ channels, each channel $z_i$ is mapped to a value in a finite set of $L$ pre-defined values, resulting in a quantized representation $\hat{z}$, which is one of $L^d$ possible vectors. An example is shown in Fig.~\ref{fig:fsq}, where $d$=3 and $L$=5, representing an implicit codebook with size $L^d=125$. Notably, when $L$ is set to $2$, each $z_i$ can take one of two possible values, yielding binary latents. This mechanism corresponds to the Lookup-Free Quantization (LFQ) method proposed in MAGVIT-v2~\citep{yu2024language}.

The experiments in Sec.~\ref{sec:abla_fsq} show that FSQ has significant advantages in codebook utilization, reconstruction quality and training stability, functioning as an advanced quantization technique that effectively improves discrete tokenizers.

\subsection{Improved Training Strategies}
\label{sec:training_strategy}

Training video tokenizers is often computationally intensive, requiring substantial resources (e.g., $3,072$ GPU hours for $256\times256$ resolution videos). This necessitates the development of efficient strategies to reduce computational costs while maintaining model performance. In this work, we implement a two-stage training approach to address this challenge: the full model is initially pre-trained on low-resolution videos, followed by fine-tuning only the decoder on high-resolution videos. Specifically, the model is first trained from scratch using videos at $128\times128$ resolution. In the second stage, the decoder is fine-tuned using videos at $256\times256$ resolution.

The experimental results presented in Tab.~\ref{tab:abla_strategy} demonstrate that the proposed two-stage training strategy achieves performance comparable to training the model from scratch on $256\times256$ resolution videos, while substantially reducing computational costs—cutting training time by half, from $3,072$ GPU hours to $1,536$ GPU hours. Furthermore, since the encoder remains unchanged, the fine-tuned model retains compatibility with the latent space of the pre-fine-tuned model. This ensures that the model can adapt efficiently to novel domains without impacting the integrity of models trained on the same latent space.

Moreover, as the video tokenizer is designed to model the motion dynamics of input videos, it is essential to efficiently represent these dynamics within the model. In this study, we empirically observe that training with data at reduced frame rates significantly enhances the model's capability to capture and represent motion dynamics. This finding is substantiated through the experimental results presented in Tab.~\ref{tab:abla_strategy} and Fig.~\ref{fig:fps8_fps3}, which illustrate the improved reconstruction quality achieved with lower frame rate training data.

\section{Experiments}
\label{sec:experiment}
This section verifies the proposed VidTok through comparative experiments with existing state-of-the-art video tokenizers~\citep{yu2024language,wang2024omnitokenizer,repocosmos,zhao2024cv,repoopensora,repoopensoraplan,yang2024cogvideox} and comprehensive ablation studies.
Fig.~\ref{fig:demo} provides several radar charts for a quick comparison.

\subsection{Experimental Setting}

\paragraph{Dataset and metrics.}
For training, we utilize a self-collected video dataset, divided into two subsets based on video quality: (1) Training Set 1, comprising approximately $10$ million low-resolution videos (e.g., 480p); and (2) Training Set 2, consisting of approximately $6$ million high-resolution videos (e.g., 1080p). All videos in the dataset are natural videos characterized by diverse lighting conditions, motion patterns, and scenarios. For evaluation, we follow the protocol of MAGVIT-v2~\citep{yu2024language} and use two benchmark datasets: the MCL-JCV dataset~\citep{wang2016mcl} and the validation set of a web video dataset. Evaluation videos are resized to $256\times256$ with a frame rate of $30$ FPS.

The video reconstruction performance of the models is assessed using four widely-used metrics: Peak Signal-to-Noise Ratio (PSNR)~\citep{hore2010image}, Structural Similarity Index Measure (SSIM)~\citep{wang2004image}, Learned Perceptual Image Patch Similarity (LPIPS)~\citep{zhang2018unreasonable} and Fréchet Video Distance (FVD)~\citep{unterthiner2018towards}.

\paragraph{Implementation details.}
We implement video tokenizers with various settings, including both causal and non-causal cases, continuous and discrete latents, and different video compression ratios.
All models are trained with four loss terms: a reconstruction term, a perceptual term, an adversarial term and a regularization term. The first three terms follow the practice in Latent Diffusion Models~\citep{rombach2022high}. For the regularization term, we use KL loss~\citep{Kingma2014} in continuous tokenizers, and entropy penalty and commitment losses in discrete tokenizers~\citep{yu2024language}.

In the first training stage, Training Set 1 is resized to a resolution of $128\times128$ and used for initial model training. We train VidTok for $50,000$ steps with batch size $16$. In the second stage, Training Set 2 is resized to $256\times256$ and employed for fine-tuning. We fine-tune the decoder for another $30,000$ steps with batch size $8$. The frame rate of the training data is maintained at $3$ frames per second (FPS) during both stages. We use Adam optimizer~\citep{kingma2014adam} with a constant learning rate of $1 \times 10^{-5}$. The training is conducted on $8$ NVIDIA 40G A100 GPUs with PyTorch~\citep{paszke2019pytorch}.

\paragraph{Baselines.}
We compare our method with the following state-of-the-art solutions:
(1) MAGVIT-v2~\citep{yu2024language}: a discrete video tokenizer which maps videos to a discrete latent space using the LFQ representation;
(2) OmniTokenizer~\citep{wang2024omnitokenizer}: a discrete video tokenizer using VQ as the discrete representation;
(3) CV-VAE~\citep{zhao2024cv}: a continuous video tokenizer with a latent space that aligns with the latent space of existing image VAEs;
(4) Open-Sora-v1.2~\citep{repoopensora}: an open-source project aimed at reproducing OpenAI’s Sora which offers a continuous video tokenizer;
(5) Open-Sora-Plan-v1.2~\citep{repoopensoraplan}: another open-source project aimed at reproducing OpenAI’s Sora;
(6) CogVideoX~\citep{yang2024cogvideox}: a continuous tokenizer that preserves a greater amount of information by maintaining a larger number of latent channels;
(7) Cosmos-Tokenizer~\citep{repocosmos}: a suite of continuous and discrete video tokenizers with various compression ratios.
We conduct thorough experiments in Sec.~\ref{sec:comparison_with_baselines}, with aligned settings for all methods to guarantee fairness in comparison.

\begin{table}[t]
\begin{center}
\footnotesize
\caption{Quantitative comparison with the state-of-the-art video tokenizers.
All evaluated models are causal and have a video compression ratio of $4\times 8\times 8$.
The input resolution for most models is $17 \times 256 \times 256$, except for MAGVIT-v2$^*$, which is evaluated on $17 \times 360 \times 640$ as reported in the original study.
The sample rate of testing data is 30 FPS.
We highlight the best and the second-best numbers in \textbf{bold} and \underline{underline} respectively.
}
    \setlength\tabcolsep{3pt}
    \renewcommand\arraystretch{1.1}
    \begin{tabular}{l|l|c|cccc|cccc}
    \toprule[1.5pt]
    \multirow{2}{*}{Method} & \multirow{2}{*}{Regularizer}  & \multirow{2}{*}{Param.}  & \multicolumn{4}{c}{MCL-JCV} & \multicolumn{4}{c}{WebVid-Val} \\
    \cmidrule(l){4-7}
    \cmidrule(l){8-11}
     &   &  & PSNR$\uparrow$ & SSIM$\uparrow$ & LPIPS$\downarrow$ & FVD$\downarrow$ & PSNR$\uparrow$ & SSIM$\uparrow$ & LPIPS$\downarrow$ & FVD$\downarrow$ \\
    \midrule 
    MAGVIT-v2$^*$  & LFQ - $262,144$   & - & 26.18 & - & 0.104 &-&-&- &-&- \\
    OmniTokenizer & VQ \, - $8,192$ & 51M & 26.93 & 0.841 & 0.165 & 232.7 & 26.26 & \underline{0.883}& 0.112 & 48.46 \\
    Cosmos-DV & FSQ - $64,000$ & 101M & 28.07 & 0.743 & 0.212 &227.7 & 29.39 & 0.741& 0.170& 57.97\\
    Ours-FSQ & FSQ - $32,768$ &  157M & \underline{29.16} & \underline{0.854} & \underline{0.117} & \underline{196.9} & \underline{31.04} & \underline{0.883} & \underline{0.089} & \underline{45.34} \\
    Ours-FSQ & FSQ - $262,144$ & 157M& \textbf{29.82} & \textbf{0.867} & \textbf{0.106} & \textbf{160.1} & \textbf{31.76} & \textbf{0.896} & \textbf{0.080} & \textbf{38.17} \\   
    \midrule
    CV-VAE  & KL - $4$chn  &  182M & 28.56 & 0.823 & 0.163 & 334.2& 30.79 & 0.863& 0.116& 70.39\\
      Open-Sora-v1.2  & KL - $4$chn  & 393M  & \underline{29.44} & 0.766 & 0.164 & 350.7 & \underline{31.02} & 0.764 & 0.137 & 112.34 \\
    Open-Sora-Plan-v1.2  & KL - $4$chn & 239M & 29.07 & \underline{0.839} & \underline{0.131} & \underline{201.7} & 30.85 & \underline{0.869} & \underline{0.101} & \underline{44.76} \\
    Ours-KL  & KL - $4$chn & 157M  & \textbf{29.64} & \textbf{0.852}& \textbf{0.114}& \textbf{194.2} & \textbf{31.53} & \textbf{0.878} & \textbf{0.087} & \textbf{36.88} \\
    \midrule
    CogVideoX & KL - $16$chn & 206M  & \underline{33.76} & \underline{0.930} & \underline{0.076} & \underline{93.2} & \underline{36.22}& \underline{0.952}& \underline{0.049}& \underline{15.30}\\
    Cosmos-CV & AE - $16$chn & 101M  & 31.27 & 0.817 & 0.149 & 153.7& 33.04 &0.818& 0.107& 23.85\\
    Ours-KL & KL - $16$chn & 157M & \textbf{35.04} & \textbf{0.942}& \textbf{0.047}& \textbf{78.9}& \textbf{37.53} & \textbf{0.961} & \textbf{0.032} & \textbf{9.12} \\
    \bottomrule[1.5pt]
    \end{tabular}
    \label{tab:general_comp}
\end{center}
\end{table}

\begin{figure}[!h]
  \centering
  \includegraphics[width=0.95\linewidth]{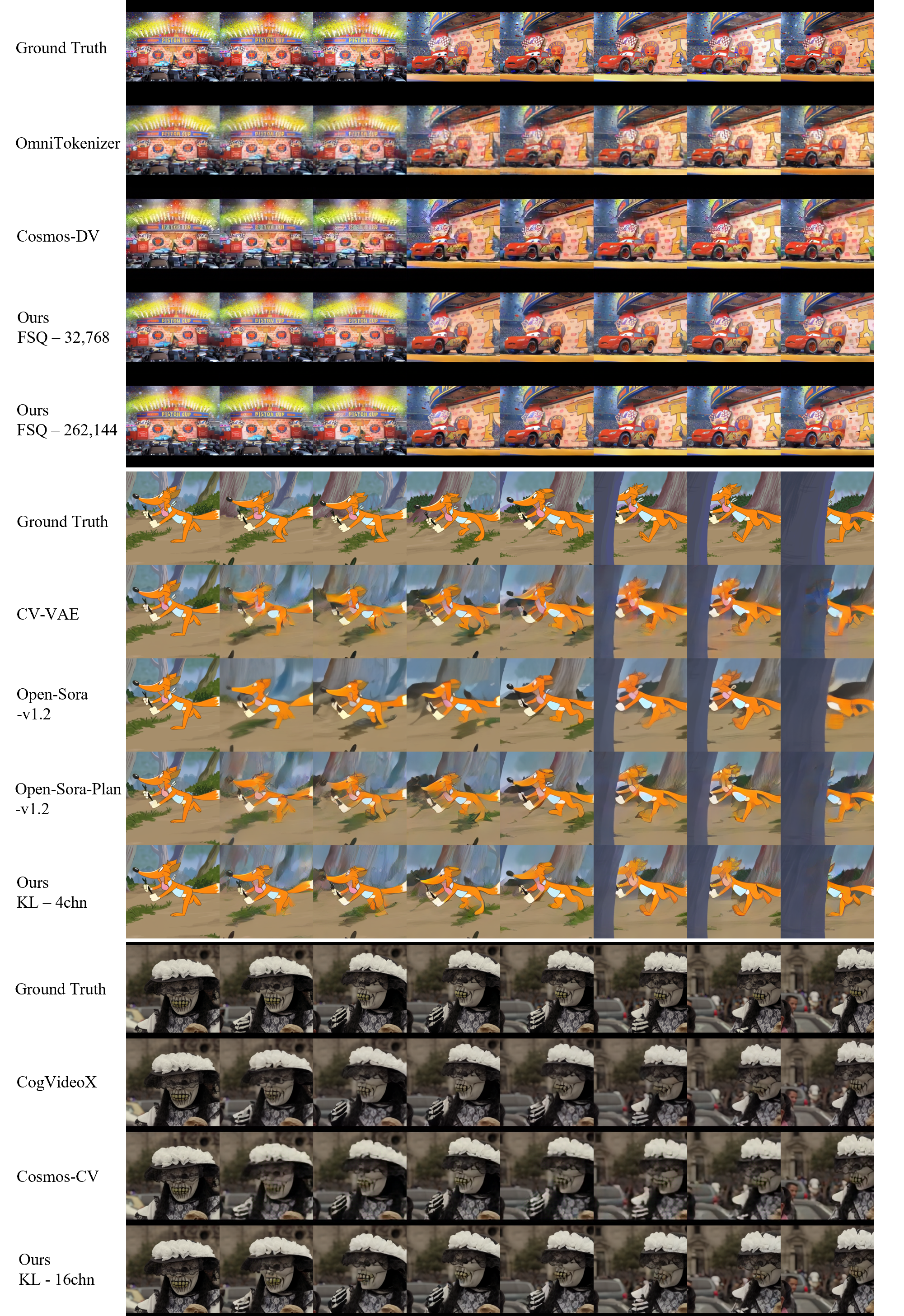}
  \vspace{-3mm}
  \caption{Qualitative comparison with the state-of-the-art video tokenizers.}
  \label{fig:full_comp}
\end{figure}

\subsection{Comparison with Baselines}
\label{sec:comparison_with_baselines}

To evaluate the advancements achieved by VidTok, we compare its performance against state-of-the-art models across various scenarios, encompassing both discrete and continuous tokenization approaches. The comprehensive comparison results are presented in Tab.~\ref{tab:general_comp}. All performance metrics reported in the table, except for those of MAGVIT-v2~\citep{yu2024language}, are obtained through our own experiments conducted under an identical evaluation protocol to ensure consistency and fairness. For MAGVIT-v2, as the model is not publicly accessible, we reference the results reported in their original publication. It is important to note that these results were obtained on a resolution of $17 \times 360 \times 640$, differing from the $17 \times 256 \times 256$ resolution used for the other models in our comparison.

Compared to existing discrete tokenziers~\citep{yu2024language,wang2024omnitokenizer,repocosmos}, VidTok demonstrates significantly superior reconstruction performance, even when utilizing a smaller codebook size (e.g., $32,768$). This highlights the effectiveness of our approach in discrete tokenization. In the context of continuous tokenization, VidTok achieves comprehensive improvements across all evaluation metrics, regardless of whether the latent representation comprises $4$ or $16$ channels. Notably, these advancements are achieved even with a smaller model size, surpassing the performance of state-of-the-art methods~\citep{zhao2024cv,repoopensora,repoopensoraplan,yang2024cogvideox,repocosmos}. These results underscore the effectiveness of VidTok in both discrete and continuous tokenization tasks.

We present the corresponding visual reconstruction results in Fig.~\ref{fig:full_comp} for qualitative comparison. From these visual results, our method exhibits a distinct advantage in detail reconstruction fidelity and subjective viewing experience.

\subsection{Ablation Experiments}

We conduct comprehensive ablation experiments to validate the superiority of the proposed model architecture, the advanced quantization technique and the improved training strategies. All ablation experiments are conducted with a video compression ratio of $4\times 8\times 8$ and an input size of $17\times 256 \times 256$, evaluated on MCL-JCV~\citep{wang2016mcl}.

\subsubsection{Ablation on the Model Architecture}

To evaluate the effectiveness of our proposed model architecture, we compare it with three alternative variants in terms of computational complexity and reconstruction quality. (1) \textbf{Variant 1} employs a fully 3D architecture, integrating spatial and temporal sampling using 3D convolutions. (2) \textbf{Variant 2} separates spatial and temporal sampling, but does not incorporate the AlphaBlender operator for temporal sampling. (3) \textbf{Variant 3} replaces all 3D convolutions with 2D convolutions.

The experimental results, summarized in Tab.~\ref{tab:abla_arch}, provide insights into the trade-offs between model performance and computational efficiency. The results indicate that employing a fully 3D architecture (Variant 1) results in high computational complexity and model size. By modifying the architecture to replace 3D convolutions in the spatio-temporal sampling modules with a combination of 2D and 1D convolutions (Variant 2), we achieve a significant reduction in computational load without notable degradation in reconstruction quality. Building upon Variant 2, the introduction of the AlphaBlender operator for temporal sampling yields substantial improvements across most metrics, albeit with a slight increase in computational cost. Furthermore, replacing all 3D convolutions with 2D convolutions (Variant 3) leads to a marked decline in reconstruction performance, underscoring the importance of retaining 3D convolutions for effective spatio-temporal representation. Overall, the findings in Tab.~\ref{tab:abla_arch} highlight the efficacy of the proposed architecture, which strikes a balance between computational efficiency and reconstruction performance.

\begin{table}[t]
\begin{center}
\small
\caption{Ablation study on the model architecture. Variant 1: fully 3D architecture. Variant 2: w/o AlphaBlender. Variant 3: w/o 3D architecture. We use `KL - 4chn' as regularizer for all settings.}
	\label{tab:abla_arch}
        \setlength\tabcolsep{12pt}
	\begin{tabular}{l|cc|cccc}
		\toprule[1.5pt]
		Method  & Param. & FLOPs & PSNR$\uparrow$ & SSIM$\uparrow$ & LPIPS$\downarrow$  & FVD$\downarrow$ \\
		\midrule
		Variant 1 & 245M & 16.98 T& 29.39 & 0.847 & 0.117 & 176.9 \\
		Variant 2 & 142M & 7.17 T& 29.36 & 0.846 & 0.119  & 185.7 \\
		Variant 3 & 126M & 10.18 T& 29.26  & 0.846 &  0.120 & 200.6 \\
		  Ours & 157M & 10.35 T & 29.64 & 0.852 & 0.114 & 194.2  \\
		\bottomrule[1.5pt]
	\end{tabular}
 \vspace{-3mm}
\end{center}
\end{table}

\subsubsection{Ablation on the Discrete Techniques}
\label{sec:abla_fsq}

\begin{table}[t]
\begin{center}
\footnotesize
\caption{Analysis of the impact of discrete techniques on model performance. R.L. denotes Regularization Loss, while U.R. represents Utilization Rate.}
	\label{tab:abla_fsq}
        \setlength\tabcolsep{8pt}
        \renewcommand\arraystretch{1.1}
	\begin{tabular}{lc|ccccc}
		\toprule[1.5pt]
		Regularizer & w/ R.L. & PSNR$\uparrow$ & SSIM$\uparrow$ & LPIPS$\downarrow$ & FVD$\downarrow$ & U.R.$\uparrow$\\
		\midrule
		  VQ - 262,144 & \usym{2613} & - & - & - & - & -\\
		  VQ - 262,144 & \checkmark & 23.22 & 0.657 & 0.336 & 960.5 & 0.2\%\\
          \midrule
		LFQ - 262,144 & \usym{2613} & 23.91 & 0.688 & 0.251 & 619.8 & 4.2\% \\
		LFQ - 262,144 & \checkmark & 28.04 & 0.833 & 0.133 &208.1 & 99.9\% \\ 
          \midrule
		FSQ - 262,144 & \usym{2613} & 29.75 & 0.866 & 0.109 &  167.5 & 99.8\% \\ 
		FSQ - 262,144 & \checkmark & 29.82 & 0.867 & 0.106 & 160.1 & 99.8\% \\
  FSQ - 32,768 & \checkmark & 29.16 & 0.854 & 0.117 & 196.9 & 100.0\%\\
  FSQ - 4,096 & \checkmark & 28.36 & 0.832 & 0.133 & 218.1 &  100.0\%\\
		\bottomrule[1.5pt]
	\end{tabular}
\end{center}
\end{table}

In Tab.~\ref{tab:abla_fsq}, we present a comparison of various quantization methods, including VQ~\citep{van2017neuralvqvae}, LFQ~\citep{yu2024language}, and FSQ~\citep{mentzer2024finite}. Additionally, we analyze the impact of the regularization loss term on the performance of discrete tokenizers.

The results highlight several key observations. Traditional VQ suffers from common challenges, such as training instability and codebook collapse, which lead to extremely low codebook utilization and suboptimal reconstruction quality. In contrast, LFQ and FSQ achieve nearly 100\% codebook utilization by directly optimizing an implicit codebook, resulting in significantly enhanced tokenizer performance. Furthermore, FSQ outperforms LFQ's binary quantization by achieving better reconstruction fidelity, suggesting reduced information loss during the quantization process.

The effects of regularization loss vary across quantization methods. For conventional VQ, the absence of regularization loss leads to model collapse and convergence failure. In the case of LFQ, while the model remains capable of convergence without regularization, it experiences a marked decline in codebook utilization and reconstruction performance. FSQ, on the other hand, demonstrates superior training stability, with its performance remaining largely unaffected even in the absence of the regularization loss term.

In summary, FSQ emerges as a highly effective quantization technique, offering significant advantages in codebook utilization, reconstruction quality, and training stability. These attributes position FSQ as an advanced method for enhancing the performance of discrete tokenizers.

\subsubsection{Ablation on the Training Strategies}

\begin{table*}[t]
\begin{center}
\small
\caption{Ablation study on the proposed training strategy. To ensure a fair comparison, both stages use training data from Training Set 1. Across all configurations, the regularizer `KL - 4chn' is employed. The training computational cost, measured in GPU hours, is evaluated using NVIDIA A100 GPUs.}
	\label{tab:abla_strategy}
        \setlength\tabcolsep{4pt}
        \renewcommand\arraystretch{1.1}
	\begin{tabular}{l|ccc|cccc|c}
		\toprule[1.5pt]
		Sample Rate & First Stage & Second Stage & Fix Enc. & PSNR$\uparrow$ & SSIM$\uparrow$ & LPIPS$\downarrow$ & FVD$\downarrow$ & GPU Hours\\
		\midrule
		  3 FPS& $256 \times 256$ & - & - & 29.19 & 0.843 & 0.127 & 174.9 & 3,072\\
		  3 FPS & $128 \times 128$ & - & - & 29.02  & 0.838 & 0.130 & 221.7 & 960\\
		  3 FPS & $128 \times 128$ & $256 \times 256$ & \usym{2613} & 29.15 & 0.842 & 0.127 & 203.2& 1,728\\
		3 FPS & $128 \times 128$ & $256 \times 256$ & \checkmark & 29.21& 0.843& 0.125 & 189.8 &  1,536 \\
		8 FPS & $128 \times 128$ & $256 \times 256$ & \checkmark & 29.02 & 0.839 & 0.126 & 219.2 & 1,536\\
		\bottomrule[1.5pt]
	\end{tabular}
 \vspace{-3mm}
\end{center}
\end{table*}

As detailed in Sec.~\ref{sec:training_strategy}, we employ a two-stage training strategy: pre-training the full model on low-resolution videos, followed by fine-tuning only the decoder on high-resolution videos. To evaluate the efficiency and effectiveness of this approach, we conduct an ablation study, with results summarized in Tab.~\ref{tab:abla_strategy}.

In the first row, training the full model on high-resolution videos directly from scratch requires 3,072 GPU hours. In contrast, the results in the fourth row demonstrate that the proposed two-stage training strategy—starting with low-resolution data and then fine-tuning on high-resolution data—reduces training time by half (from 3,072 to 1,536 GPU hours) while achieving comparable reconstruction quality. A comparison between the third and fourth rows reveals that fine-tuning only the decoder during the second stage produces similar performance to fine-tuning the entire model, with a lower computational cost. This approach also ensures that the low-resolution and high-resolution models share a unified latent space due to the fixed encoder, enabling latent models trained in this shared space to be reused across resolutions and domains.

\begin{figure*}[t]
  \centering
  \includegraphics[width=0.99\linewidth]{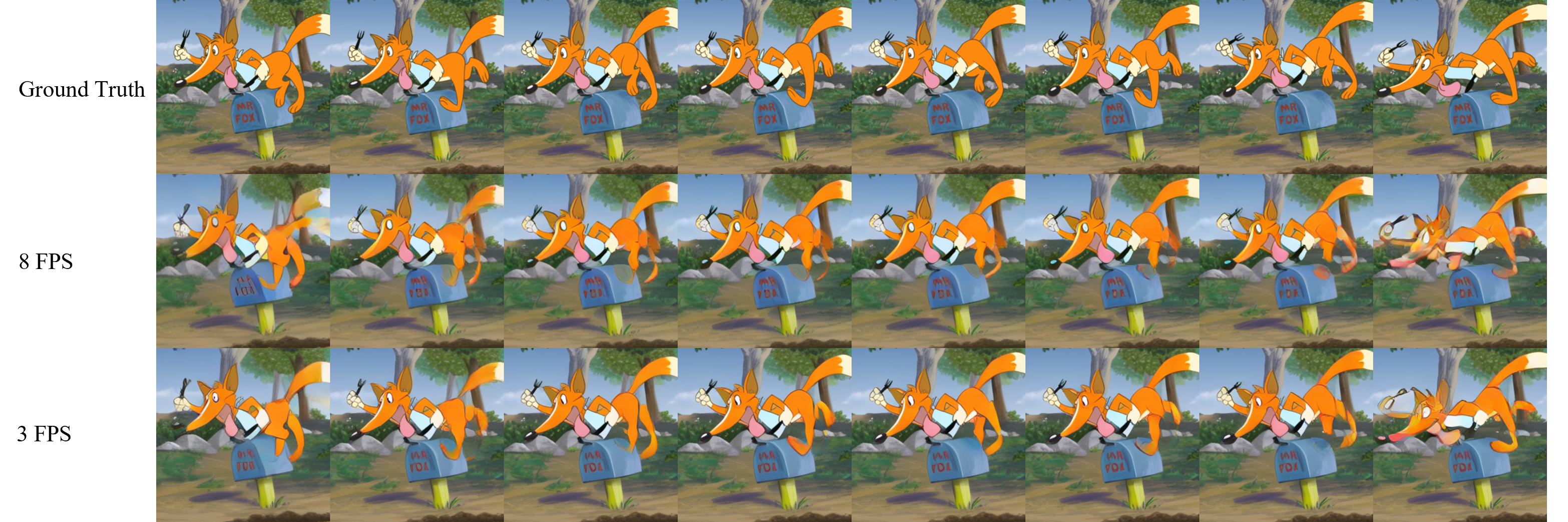}
  \caption{The influence of different sample rates on model performance during training. The second row presents the test results obtained using training data with a sample rate of 8 FPS, while the third row shows the test results using training data with a sample rate of 3 FPS. The results demonstrate that employing training data with reduced frame rates enhances the model's capacity to effectively capture motion dynamics.}
  \label{fig:fps8_fps3}
\end{figure*}

Additionally, the last row examines the impact of varying the sampling rate during training. Qualitative results, presented in Fig.~\ref{fig:fps8_fps3}, indicate that using training data with reduced frame rates enhances the model’s ability to represent motion dynamics effectively.

\subsection{Model Summary}

\begin{table*}[t]
\begin{center}
\small
\caption{Model summary. We offer a suite of models with diverse configurations, encompassing both continuous and discrete tokenization, various video compression ratios (VCR), and options for causal and non-causal scenarios. These configurations are designed to address the distinct requirements of various downstream tasks.}
	\label{tab:more_results}
        \setlength\tabcolsep{4pt}
        \renewcommand\arraystretch{1.1}
	\begin{tabular}{lcccc|c|cccc}
		\toprule[1.5pt]
		Regularizer  & Causal & Input Size & VCR & Latent Size & Param. & PSNR$\uparrow$ & SSIM$\uparrow$ & LPIPS$\downarrow$ & FVD$\downarrow$\\
		\midrule
		KL - 4chn & \checkmark & $17\times 256 \times 256$ & $4\times 8 \times 8$ & $5\times 32 \times 32$ & 157M  & 29.64 & 0.852 & 0.114 & 194.2 \\
		KL - 4chn & \checkmark & $17\times 256 \times 256$ & $4\times 16 \times 16$ & $5\times 16 \times 16$ & 199M & 25.05& 0.711& 0.228 & 549.1 \\
		KL - 4chn & \usym{2613} & $16\times 256 \times 256$ & $4\times 8 \times 8$  & $4\times 32 \times 32$ & 158M&30.60 &0.876 & 0.098&157.9\\
		KL - 4chn & \usym{2613} & $16\times 256 \times 256$ &  $4\times 16 \times 16$& $4\times 16 \times 16$ & 199M& 26.06 & 0.751 & 0.190 & 423.2\\
        \midrule
  KL - 8chn & \checkmark &$17\times 256 \times 256$ & $4\times 8 \times 8$  & $5\times 32 \times 32$ & 157M& 31.83 & 0.897 & 0.083 & 109.3 \\
  KL - 16chn & \checkmark &$17\times 256 \times 256$ & $4\times 8 \times 8$  & $5\times 32 \times 32$ & 157M& 35.04 & 0.942 & 0.047 & 78.9 \\
  KL - 8chn & \checkmark &$17\times 256 \times 256$ & $2\times 8 \times 8$  & $9\times 32 \times 32$ & 149M & 33.86 & 0.928 & 0.057 & 80.7 \\
  KL - 4chn & \checkmark &$17\times 256 \times 256$ & $4\times 4 \times 4$  & $5\times 64 \times 64$ & 155M & 34.78 &0.941&0.051& 87.2\\
        \midrule
        FSQ - 4,096 & \checkmark & $17\times 256 \times 256$ &$4\times 8 \times 8$  & $5\times 32 \times 32$ & 157M & 28.36 & 0.832 & 0.133 & 218.1\\
        FSQ - 32,768 & \checkmark & $17\times 256 \times 256$ &$4\times 8 \times 8$  & $5\times 32 \times 32$ & 157M & 29.16 & 0.854 & 0.117 & 196.9 \\
        FSQ - 262,144 & \checkmark & $17\times 256 \times 256$ &$4\times 8 \times 8$  & $5\times 32 \times 32$ & 157M & 29.82 & 0.867 & 0.106 & 160.1\\
		FSQ - 262,144 & \checkmark & $17\times 256 \times 256$ & $4\times 16 \times 16$ & $5\times 16 \times 16$ & 199M& 25.38& 0.738& 0.206& 430.1\\
  FSQ - 262,144 & \usym{2613} &$16\times 256 \times 256$  & $4\times 8 \times 8$  & $4\times 32 \times 32$& 157M& 30.78& 0.889 & 0.091& 132.1\\
		FSQ - 262,144 & \usym{2613} & $16\times 256 \times 256$ & $4\times 16 \times 16$ & $4\times 16 \times 16$ & 199M& 26.37 & 0.772& 0.171& 357.0\\
		\bottomrule[1.5pt]
	\end{tabular}
\end{center}
\end{table*}

We provide a comprehensive summary of model performance in Tab.~\ref{tab:more_results}, covering both continuous and discrete tokenization, various video compression ratios, and causal versus non-causal scenarios.

From Tab.~\ref{tab:more_results}, it is evident that as the video compression ratio increases, reconstruction performance deteriorates. Non-causal models generally outperform causal ones, as they are able to capture more extensive temporal information, which aids in the high-fidelity reconstruction of fine details. In the continuous case, increasing the number of channels in the latent representation allows for the retention of more information, leading to better reconstruction performance. Similarly, in the discrete case, a larger codebook size usually means smaller quantization errors, preserving more accurate information and thus achieving better reconstruction fidelity. A comparison between the continuous and discrete cases reveals that FSQ with a codebook size of $262,144$ achieves reconstruction performance comparable to `KL - $4$chn'.

Additionally, we compare the performance across different settings: 1) KL - 16chn, with a video compression ratio of $4\times8\times8$; 2) KL - 8chn, with a video compression ratio is $2\times8\times8$; 3) KL - 4chn, with a video compression ratio is $4\times4\times4$. Our analysis indicates that when the latent space contains the same amount of data, allocating it to the channel dimension tends to result in relatively better reconstruction performance.

All models and source code associated with this work are publicly available at~\url{https://github.com/microsoft/VidTok}. We aspire for this contribution to serve as a foundation for and inspire further advancements in this research domain.

\section{Conclusion}
\label{sec:conclusion}

In this paper, we present VidTok, a versatile and open-source video tokenizer that achieves state-of-the-art performance in both continuous and discrete tokenization. By converting raw visual data into compact latent tokens, VidTok provides an efficient foundation for tasks related to visual generation and understanding. Through the incorporation of advancements in model architecture, discrete representation, and training strategies, VidTok surpasses existing methods, demonstrating notable improvements across several performance metrics, including PSNR, SSIM, LPIPS, and FVD, under standardized evaluation protocols. Additionally, we conduct extensive ablation experiments to thoroughly investigate the performance characteristics of the video tokenizer. We hope this work will inspire future research in this area.

\section*{Acknowledgment}

We extend our gratitude to all individuals who contributed their insights and efforts to this project, including Chong Luo, Ruoyu Feng, and Zhipeng Huang for their valuable discussions, and Qi Dai for his guidance on the open-source process.

\bibliography{main}
\bibliographystyle{iclr2025_conference}

\end{document}